\documentclass[11pt,a4paper]{article}
\usepackage[hyperref]{acl2020}
\usepackage{multibib,graphicx,tabularx,soul,epstopdf,hyperref,xstring}
\usepackage[latin1]{inputenc}
\usepackage{times}
\usepackage{latexsym}

\usepackage{microtype}

\aclfinalcopy 

\title{Application of the interactive Leipzig Corpus Miner as a generic research platform for the use in the social sciences}

\author{
Christian Kahmann$^{\ast}$, Andreas Niekler$^{\ast}$, Gregor Wiedemann$^{\dagger}$\\

 $^{\ast}$Natural Language Processing Department - University Leipzig \\ Augustusplatz 10, 04109 Leipzig \\ 
          \texttt{\{kahmann, aniekler\}@informatik.uni-leipzig.de} \\ \\
$^{\dagger}$Leibniz-Institut f\"ur Medienforschung - Hans-Bredow-Institut (HBI) \\ Rothenbaumchaussee 36,
20148 Hamburg \\ 
          \texttt{g.wiedemann@leibniz-hbi.de}
}

\date{}

\begin{document}
\maketitle
\begin{abstract}
This  article  introduces  to  the \textit{interactive Leipzig Corpus Miner} (iLCM) - a  newly  released,  open-source  software to perform automatic content analysis. Since the iLCM is based on the R-programming language, its generic text mining  procedures  provided  via  a  user-friendly  graphical user interface (GUI) can easily be extended using the integrated IDE RStudio-Server or numerous other interfaces in the tool. Furthermore, the iLCM offers various possibilities to use quantitative and qualitative research approaches in combination. Some of these possibilities will be presented in more detail in the following.
\end{abstract}

\section{Introduction}
The use of computational methods is becoming increasingly important in the social sciences and in its subdisciplines, such as communication science \cite{doi:10.1080/19312458.2018.1458084}. This is mainly due to the rapidly growing amount of digital data. Especially for large textual datasets, automatic procedures are required since conventional content analysis involving steps of manual reading and interpretation is not feasible any longer. Automatic approaches to content analysis in textual data promise a much more efficient processing and allow for scaling with the constantly growing amount of available data. 
Furthermore, previous studies have shown that the application of automatic methods can lead to novel insights that would not have been possible to obtain with traditional methods alone \cite{wiedemann_opening_up}.

For researchers in the applying fields, coding their own analysis programs often is not a viable option due to lack of resources or expertise.\footnote{Actually, more and more computational social scientists obtain coding skills. However, the development of complex research software remains a complicated process which usually requires trained software developers.}
Instead, applied research usually relies on existing research software. 
On the one hand, this could be standalone software solutions designed for very specific analysis purposes (e.g. a word frequency analyzer). This restrains researchers to narrow study designs remaining within the limitations of the specific software. Any desired functionality deviating from this cannot be realized leading to a significant reduction in the method portfolio of the project. 
On the other hand, generic software solutions are available which provide a larger number of analysis tools and, thus, do not restrict users to a narrow methodological workflow. Instead, generic research tools enable the application of various methods which can be flexibly combined, and, therefore, be used in a wide range of projects. Moreover, this flexibility results from opportunities for adaptation and extension of the methodical approaches built into the generic software.

The \textit{interactive Leipzig Corpus Miner} (iLCM) represents such a generic software solution for the use of text mining in the social sciences and humanities. In the following, we reflect on which features allow the tool to be customizable and extendable. Furthermore, Section~\ref{union_quanti_quali} describes the possibilities of combining quantitative and qualitative research approaches in the iLCM. Based on this, Section~\ref{application} shows exemplarily which advantages result from the described features of the tool.

\section{Related Work}
Of course, the iLCM is not the only software solution available in the field of the application of text mining in the social sciences.
Several other solutions exist. These include, for example, the various QDA software solutions such as \textit{MAXQDA}\footnote{https://www.maxqda.de/}, \textit{Atlas.ti}\footnote{https://atlasti.com/} or \textit{NVivo}\footnote{https://www.nvivo.de/} These have specialised in the process of qualitative data analysis of texts in the social sciences. These processes are excellently mapped in these software solutions. However, research questions that clearly deviate from this procedure cannot be mapped. 
Other more flexible tools include \textit{RapidMiner} \cite{rapidminer} and \textit{KNIME} \cite{Berthold:2009:KKI:1656274.1656280}.  These offer a variety of different utilities, which can then be combined and put together to form a workflow. Both \textit{RapidMiner} and \textit{KNIME} offer a graphical user interface. In addition, there are command line-based libraries such as the R packages \textit{quanteda} \cite{quanteda} or \textit{polmineR} \cite{PolminerR}, which also provide numerous tools for the use of text mining in a social science context. However, these require a certain amount of prior knowledge in the use of command line editing tools. The iLCM, on the other hand, offers both a graphical user interface as well as the option of using a command line based environment. This makes it possible not only to adapt and expand the numerous analyses already available, but also to further optimise the entire application of the tool to suit one's own problem. Additionally, the iLCM offers numerous export options to support interoperability with other software solutions.

\section{iLCM as a generic research tool}
\begin{figure*}[h]
    \centering
   \includegraphics[width=\textwidth]{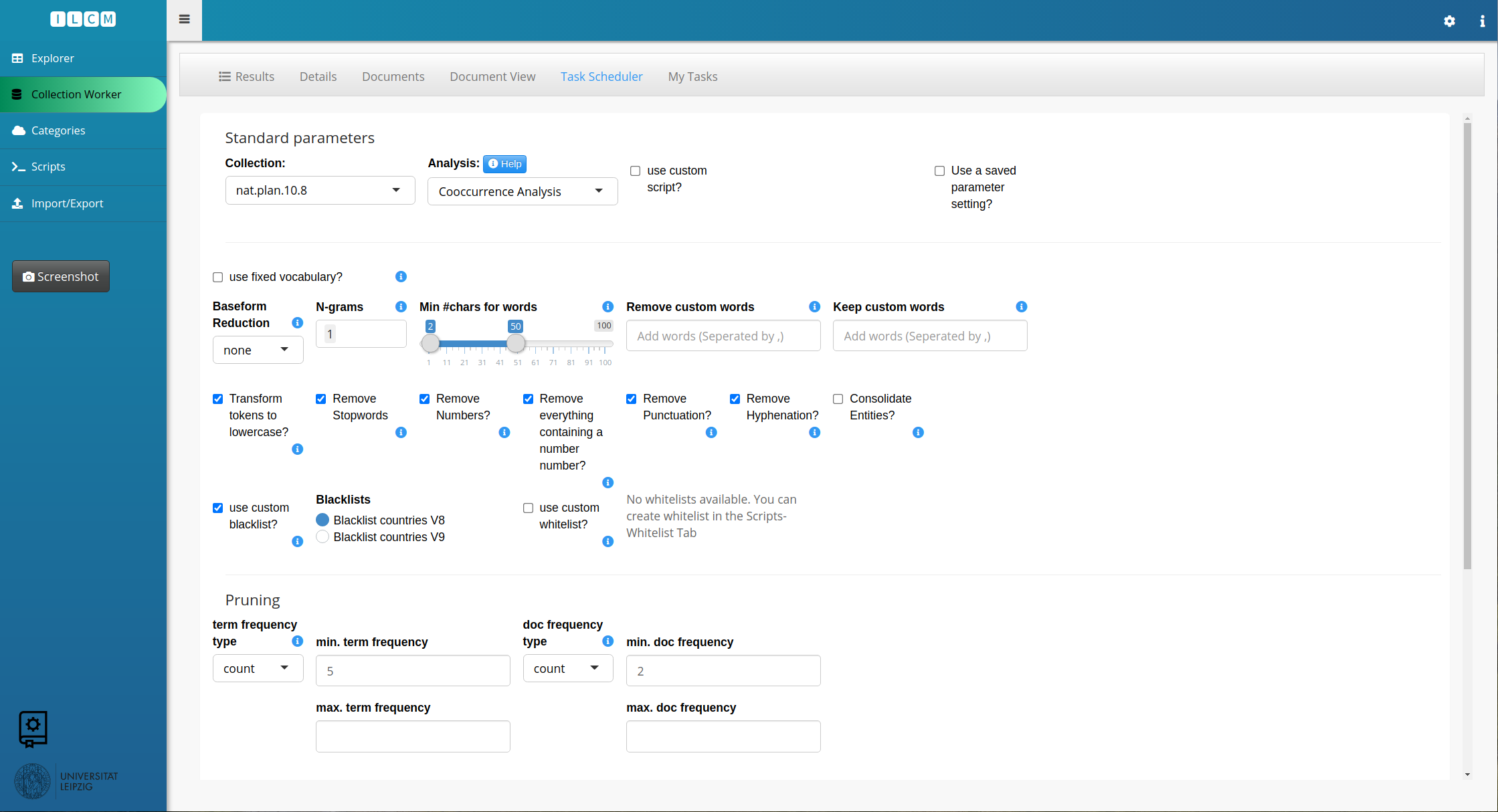}
    \caption{Interface of the iLCM for parameterisation of an analysis; In the example shown, a cooccurrence analysis is carried out for which in detail unigrams are used, words consist of at least 2 and at most 50 characters, a specially created blacklist is used, stop words and numbers are removed and pruning is also carried out. Further settings are available. The flexible parameterisation of an analysis allows different text bases (in language and quantity) as well as research questions to be processed.}
    \label{ilcm:screenshot_parameter}
\end{figure*}

For a generic research software to support automatic content analysis, specific requirements must be met. In the following, we elaborate on four functional requirements together with the solutions as implemented in the iLCM.
\paragraph{Analysis capabilities:}
Availability of a wide range of predefined functions:
In order to be able to adequately address different kinds of research questions, it is necessary to provide as many predefined functions from the areas of text mining and machine learning as possible. In the iLCM, numerous procedures are implemented to this end. The iLCM supports multi-language document pre-processing, document retrieval and collection management, content deduplication, word frequency analysis, word co-occurrence analysis, time series analysis, topic models, category coding and annotation, supervised text classification (e.g. sentiment analysis), and more. Different methods can be combined with each other in flexible ways.
\paragraph{Adaptability:}
Predefined analysis capabilities often require research specific adaptions in both, either pre-processing or analysis of textual data.
It is important to have a high tolerance for different text bases, so that various data sets, languages or metadata can be processed. In the iLCM a high degree of adaptability is guaranteed by the possibility to extensively parameterise each analysis step (see figure \ref{ilcm:screenshot_parameter}).
Since internally every analysis method is implemented as a script written in the programming language R, it is possible to adapt the predefined methods directly within the tool, and to easily add the support for further languages.

\paragraph{Extensibility:}
If some of the required functions are not available in the tool, it should be possible to add these functions. In the iLCM, new scripts can be created within the iLCM script editor to add new or replace existing analysis capabilities. For instance, it is easy to add project-specific black- and whitelists of words for pre-processing steps. Further, it is also possible to implement additional analyses based on interim results in an associated IDE.\footnote{Parallel to the iLCM GUI, an RStudio-Server instance is provided as an IDE that has access to the available data and results.} This design of the iLCM software allows researchers to apply the principles of Agile Development to carry out own implementations in a comparatively short time to realize an analysis workflow tailored to her/his individual requirements \cite{mci/Heyer2019}. In detail, here Agile Development refers to the possibilities of building on the existing infrastructures of the iLCM to be able to answer one's own questions through independent implementations. These implementations can either be carried out separately from the iLCM or integrated into it. Due to the existence of numerous functions and an already existing IDE, which can fall back on various pre-installed key libraries, initial results can be achieved very quickly in the sense of a prototypical procedure. This in turn means that necessary adjustments to the code and the operationalisation of the problem can be recognised at an early stage and correspondingly implemented in an agile manner.

\paragraph{Data export:}
If it is not possible or desired to implement a research design fully within the framework provided by the iLCM, it may still be possible to map at least partial processes with the help of the tool. The result of these sub-processes can then be exported in standard formats such as CSV or the REFI-QDA standard \cite{eversREFI} and used in other software solutions like MAXQDA or Scripting environments like R and Python.

\paragraph{Validation:}
We have consistently paid attention to validation methods during implementation. In all methods for which a standard evaluation such as Precision, Recall, F1 or Sementic Coherence is described in the literature, an evaluation routine was built in to validate the results. This allows the specification and application of quality assurance processes for scientific publications and strengthens the confidence in automatic processes for textual content analysis.

The shown features, with a variety of already existing algorithms, which can be adapted to the respective research question, as well as the possibility to add further functions to the tool, allow the use of the tool in various fields of social sciences.

\section{Combination of quantitative and qualitative approaches}
\label{union_quanti_quali}
The analysis capabilities of the iLCM allow for combinations of quantitative with rather qualitative (interpretive) research approaches to investigate large textual datasets. This combination of the two research paradigms can be achieved in different ways. In the following, we present three main hybrid research concepts and explain how they can be realized within the iLCM software.
\paragraph{Text classification:}
Qualitative Data Analysis (QDA) is a basic research method in the social sciences. Here, texts, resp. text segments are coded into different categories on the basis of a previously defined codebook. Then, distributions of codings in texts are used for further descriptive analysis. However, the manual coding of large amounts of text is time-consuming and costly. To support manual coding, computational methods such as supervised machine learning algorithms can be applied. These try to train a model which is able to predict the categories of the uncoded data based on a set of data manually coded by the researcher. To improve the performance of such a machine classifier efficiently, approaches of active learning \cite{settles2010active} come into place. 
In a mutual interplay of manual and automatic procedures, it is thus possible to obtain a qualitatively based classification of data due to the researchers interpretive decisions during training data creation. At the same time, automatic coding allows for scaling the coding process to very large datasets ready for quantitative analysis of qualitative codings.

\paragraph{Topic model validation:}
The result of a topic model analysis \cite{Blei:2003:LDA:944919.944937} is described by a set of topics that represent probability distributions over the vocabulary. The documents in turn are reflected as distributions of these same topics.
Topic models are based on the significant common occurrence of semantically related words. Ideally, this makes it possible to find word clusters that cover distinct semantic areas which can be interpreted as topics. If this is the case, further analyses can be carried out on the basis of these thematic clusters, for example in relation to existing metadata. However, the assessment of when a model captures interpretable topics sufficiently cannot be done automatically alone, but also requires qualitative steps. Knowing this, the iLCM has a built-in interface which allows the qualitative validation of topic model results. For this purpose the original documents are displayed. Depending on the selected topic, the words in these documents can be highlighted in colour to show which section of text is responsible for the assignment of a document to a certain topic (see Fig.~\ref{ilcm:validation}). The researcher has the possibility to compare the word distributions found and to check their plausibility with his domain expert knowledge. In this way, possible sources of error during the import of the data, such as the presence of duplicate files or defective OCR, can also be detected and addressed in a subsequent step.

\paragraph{Thematic filtering:}
Simply put, the result of a topic model provides clusters of semantically coherent word probabilities and the association of these clusters with documents. To assign meaning to these clusters qualitative interpretation is required. This cannot be done by interpreting the word clusters alone (cp. the process of topic model validation above), but also by reading topic representative documents. For this purpose, the iLCM provides the possibility to select texts a topic model is based on according to its topic composition. Selected texts can be viewed with coloured highlightings to visualize the thematic affiliations of words. This makes it possible to understand which text passages are typical for a certain topic and, thus, enable a better understanding of what aspects a single topic is actually composed of. Once the topics have been interpreted, they can be linked to existing metadata in the tool in order to derive results and hypotheses from the model.


\section{Exemplary study}
\label{application}

\begin{figure*}[tb]
\begin{minipage}[t]{0.46\textwidth}
\includegraphics[width=\textwidth]{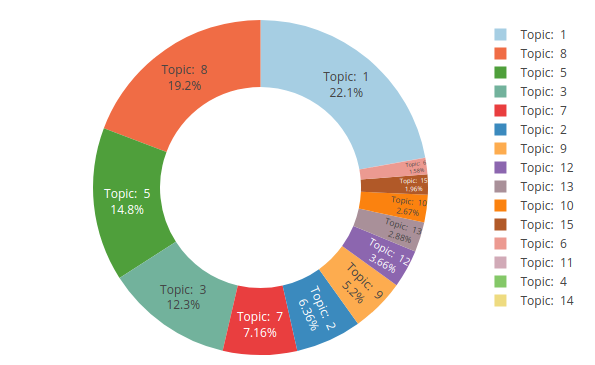}
\caption{Topic distribution of the NDC of Gambia.}
\label{fig:vola_dice_nopp}
\end{minipage}
\hfill
\begin{minipage}[t]{0.46\textwidth}
\includegraphics[width=\textwidth]{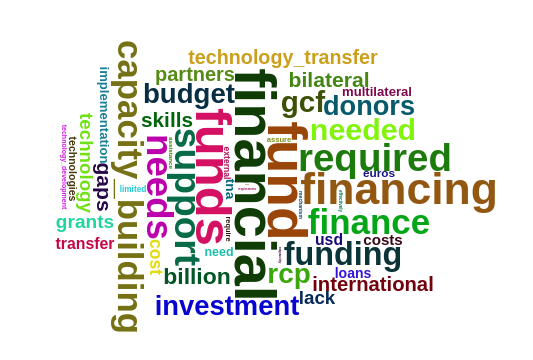}
\caption{Wordcloud to display the most relevant words for Topic 8. This Topic has been labelled as \textit{international support}.}
\label{fig:vola_dice_nopp_log}
\end{minipage}
\end{figure*}

\begin{figure*}[h]
    \centering
   \includegraphics[width=\textwidth]{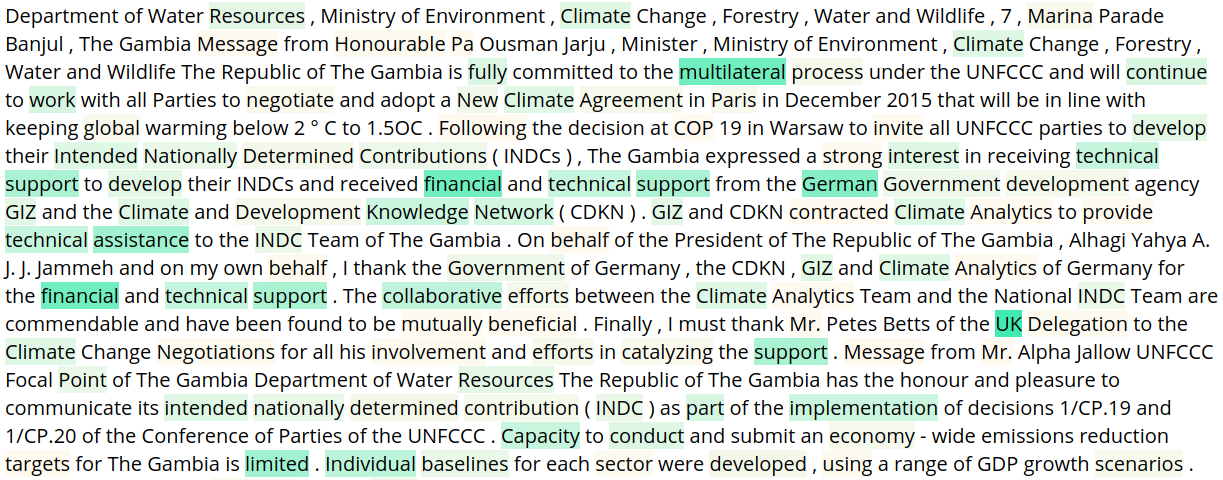}
    \caption{Validation interface in which the text of the NDC of Gambia is displayed. In addition, the most relevant words for the selected Topic 8 are highlighted in colour. The visualisation makes it very easy to see that a) this section does indeed address a funding issue and b) the exact content focus here on the financial and technical support provided by GIZ and CDKN is evident.}
    \label{ilcm:validation}
\end{figure*}

An example application within the TRANSNORMS project\footnote{https://www.transnorms.eu} demonstrates some of the described analysis capabilities of the iLCM. The TRANSNORMS project seeks to examine the translation of global (political) norms to the local level. In this study, particularly the understanding of the norms on climate protection agreed on at the Climate Conference 2015 (COP~21) in Paris are examined. At COP~21, 195 countries concluded the first comprehensive and legally binding global climate protection agreement. 
As a result, all participating countries committed themselves to submit nationally determined contributions (NDC) in which they list their intentions to achieve the jointly agreed targets. These NDCs provide the data basis for the analysis. 

For the analysis, the NDC texts were imported into the iLCM. At this point, the focus on a variable import interface was beneficial, which allows different document types such as .pdf, .doc or .csv and .xlsx to be uploaded and then interactively mapped to the data format of the iLCM in the graphical interface of the tool.
The aim of the project was then to investigate to what extent unsupervised procedures are able to identify different thematic fields within the NDCs. Based on this, it was to be further investigated whether the distribution of the texts into the thematic fields would result in correlations with country-specific metadata. To answer these questions, the LDA topic model as provided in the iLCM was used. To not receive topics governed by geographic entities, it was necessary to remove the proper location names such as countries or cities from the documents. For this purpose a blacklist was created inside the iLCM building on the already existing named entity tags\footnote{The named entity tags are assigned as part of the import of new data into the iLCM.} as a list of candidates. Based on these pre-processing steps, a final model was then calculated. The topics found were then checked for validity (see Fig.~\ref{ilcm:validation}) and subsequently interpreted. Subject domains such as: \textit{Renewable Energy, Economic growth, Water-related vulnerability} and \textit{UNFCCC collaboration} could be identified and labeled\footnote{The most relevant words of a topic as well as typical text passages (\ref{ilcm:validation}) were used to define the labels. For the topic \textit{international support}, for example, the most relevant words were: \textit{financial, fund, funds, financing, required, support, needs, finance, capacity\_building, investment, donors,...}}. It was thus established that it is possible to identify thematic areas unsupervisedly in the NDCs. The issues found in this way were then examined using metadata analysis tools added to the iLCM specifically for this purpose. Among other things, it was found that there is a clear distinction between Annex-1 and Non-Annex countries. In some topics, however, more surprising results were also found. For example, for the topic titled \textit{Water-related vulnerability}, high probability values were found for states such as island states, which are fairly obviously affected by their geographical location. Surprisingly, at the same time states that were considered a priori as rather sceptical with regard to measures for climate protection also showed high shares in this topic.
\newline
In summary, the iLCM was used here to uncover the initial questions regarding the various thematic priorities within the NDCs of the various countries and negotiating groups. A workflow was established, which will scale much better with increasing data volumes compared to purely qualitative approaches, which are rather commonly used in this field. 
For this, the possibilities of adaptability and expandability were essential to meet the specific requirements of the research question. The possibilities for qualitative assessment of the quantitative results were used. This allowed for a very efficient and in-depth evaluation, validation and interpretation of the found distribution of topics, in order to be able to make conclusive findings on the positions of the countries on the various aspects of climate change/climate protection.

\bibliography{bibliography.bib}
\bibliographystyle{acl_natbib}

\end{document}